\documentclass[10pt, a4paper]{article}
\usepackage{lrec2000}

\usepackage{times}
\usepackage{latexsym}
\usepackage{alltt}
\usepackage{graphicx}
\usepackage{array}

\title{Context-related Derivation of Word Senses}

\name{Manuela Kunze and Dietmar R{\"o}sner}

\address{ Otto-von-Guericke-Universit{\"a}t Magdeburg\\
Institut f{\"u}r Wissens- und Sprachverarbeitung \\
P.O. Box 4120, D--39016 Magdeburg,\\ Germany \\
               makunze, roesner@iws.cs.uni-magdeburg.de}

\abstract {Real applications of natural language document
processing are very often confronted with domain specific lexical
gaps during the analysis of documents of a new domain. This paper
describes an approach for the derivation of domain specific
concepts for the extension of an existing ontology. As resources,
we need an initial ontology and a partially processed corpus of a
domain. We exploit the specific characteristics of the sublanguage
in the corpus. Our approach is based on syntactic structures (noun
phrases) and compound analysis to extract information required for
the extension of GermaNet's lexical resources.}
\begin{document}
\maketitleabstract
\section{Introduction}
One of the bottlenecks in real applications of natural language
document processing is the coverage of domain-specific lexical
resources. In experiments with the document suite
XDOC\footnote{XDOC stands for \emph{X}ML based \emph{doc}ument
processing.}, we currently are processing documents about casting
technology, company profiles from web pages, and autopsy
protocols. Many of the tools have an extensive need for linguistic
resources. Therefore we are interested in ways to exploit existing
resources with a minimum of extra work. The resources of GermaNet
promise to be helpful for different tasks in the workbench. 

In this paper, we will outline how the resources of GermaNet can
be extended. Our methods exploit the specific characteristics of
the documents in the corpus. We combine different approaches to
extract new concepts from the corpus. The idea behind our approach
is to generalise from structures with known GermaNet entries to
structures without GermaNet entries.

This paper presents only experiments with GermaNet on German
texts, but the approach can also be applied on WordNet when
processing domain specific English texts.

The paper is organized as follows: The next section briefly
outlines the test corpus and the integration of GermaNet in XDOC.
Section 3 describes the methods for the extraction of new concepts
and the results. We conclude the paper with a discussion section.

\section{Document Processing with XDOC}
\subsection{Characteristics of the Corpus}\label{sec-autopsy}
In the following description of the approach, a corpus of forensic
autopsy protocols is used, because these documents are especially
amenable to processing with techniques from computational
linguistics and knowledge representation.

Autopsy protocols consist of the following major document parts:
\emph{findings}, \emph{histological} \emph{findings},
\emph{background}, \emph{discussion}, \emph{conclusions}, etc. Our
analyses focus on the sections of \emph{findings},
\emph{background} and \emph{discussion}. In the \emph{findings}
section, a high ratio of nouns and adjectives is encountered and
the sentences, which can also be verbless, are mostly short. This
section describes the medical findings in a common language. Here
we find no domain specific (medical) terms. The \emph{background}
and \emph{discussion} sections contain a standard distribution of
all word classes and regular syntactic structures. The
\emph{background} section describes, for example, the details of a
traffic accident, while the section \emph{discussion} contains a
combination of the results of the \emph{finding} section and the
facts reported in the \emph{background} section.

\subsection{Integration of GermaNet} The document suite XDOC
contains methods for linguistic processing of documents in German.
The focus
of the work has been to offer end users a collection of highly
interoperable and flexible tools for their experiments with
document collections. 

XDOC consists of different modules, for example, the syntactic
module and the semantic module (for a more detailed description
see \cite{roesner.kunze.2002coling}):

For the semantic analyses of a domain using XDOC, knowledge about
the domain -- ideally a domain specific ontology -- is needed. One
possible resource for the processing of autopsy protocols could be
medical thesauri like UMLS (Unified Medical Language
System).\footnote{http://www.nlm.nih.gov/research/umls/umlsmain.html}
Many of these resources work with medical
terminology, but in the corpus of forensic autopsy protocols only
everyday terms are used. Thus a resource that contains everyday
terms and concepts (and their relations) from the medical domain
is required for the analysis. GermaNet (see
\cite{hamp.feldweg:1997}, \cite{kunze:2001}) is intended as a
model of the German base vocabulary.


However, specific terms in some particular domains, like the
medical domain, are covered only partially in GermaNet.

For the semantic analysis in XDOC, the \emph{synonymy} and the
\emph{hypernymy} relations of GermaNet are used. We found a good
coverage of GermaNet's resources for terms in the corpus: section
\emph{findings} with 31 \%, section \emph{background} with 44 \%,
and section \emph{discussion} with 42 \% coverage (see also
\cite{kunze.roesner.2003germanet}). The reason for the poor
\emph{findings}'s result is the high frequent occurrence of
medical concepts denoted by noun compounds like \emph{Nierengewebe
(kidney tissue)} or \emph{Halswirbels\"aule (cervical spine)} that
are not covered by GermaNet, whereas the individual compound words
like kidney and spine have lexical entries in GermaNet.

In the next section, we will describe how new entries can be
derived from entries that exist in GermaNet. We start with a
corpus of autopsy protocols parsed syntactically by XDOC and with
GermaNet as an initial ontology.

\section{Methods for the Deduction of Word Senses}
In \cite{roesner02}, we outlined some ideas for the exploitation
of sublanguage characteristics of a corpus for lexicon creation.
In this paper, we will further elaborate these ideas. This section
presents how the syntactic structures of the corpus sublanguage
can be useful for the extraction of new GermaNet entries.

\subsection{Fundamental Idea of the Approach}
In the \emph{findings} section of the documents, high-frequency
complex noun phrases can be exploited for the extension of the
GermaNet resources.

The grammar fragment used in XDOC for this corpus covers the
following complex noun phrases (In all cases, the first NP is a
simple noun phrase.):
\begin{itemize}
    \item NP  NP$_{genitive}$,
    \item NP  NP$_{genitive}$ *PP, and
    \item NP *PP.
\end{itemize}

Our experiments are based on the interpretation of complex noun
phrases that are described by the syntactic structure NP
$\rightarrow$ NP NP$_{genitive}$ (i.e. a simple NP modified by a
genitive attribute).

In the case of a complex noun phrase, several possibilities for a
semantic interpretation of this syntactic structure exist, for
example, \texttt{part-of} relations in \emph{'dermis of the hand'}
or \texttt{patient-of} relation in \emph{'the production of
cars'}.

\small
\begin{figure}[hbtp]
\begin{center}
\includegraphics[scale=0.55]{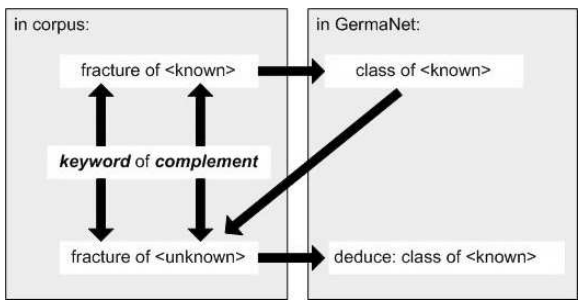}
\caption{A Sketch of the Idea.}\label{idea}
\end{center}
\end{figure}

\normalsize

The idea behind the approach is based on following assumptions. A
structure of the form \emph{KEYWORD of COMPLEMENT} describes the
same relation for every possible candidate of the complement,
e.g., \texttt{part-of}. Further on, an assumption is that the
complement candidates of a keyword have the same semantic
category. The information of complement candidates available in
GermaNet is used to deduce information about the semantic category
of candidates that are unknown in GermaNet (see also Fig.
\ref{idea}).

\subsection{Exploiting Syntactic Structures of the Corpus}

In the corpus (of 600 autopsy protocols and more than 1.5 million
word forms), structures in the form of \sloppypar \texttt{NP
$\rightarrow$ NP NP$_{genitive}$} are often encountered. For
example, the phrase \emph{'Schleimhaut des Magens'} (\emph{mucosa
of the stomach}) occurs 317 times in the corpus. The more
generalised phrase \emph{'mucosa of XXX'} occurs 836 times in the
corpus. Another generalised example is the phrase \emph{'fracture
of XXX'} that occurs 749 times in 93 different forms. One example
form is the class of NPs with keyword 'Bruch' (fracture) and
modified by a complement (the second noun phrase in the
structure), e.g.,
\emph{'Wirbels\"aule' (spine)} in the phrase \emph{'Bruch der
Wirbels\"aule'} (occurs 58 times) or \emph{'Wadenbein' (fibula)}
in the phrase \emph{'Bruch des Wadenbeines'} (occurs 11 times).
Other complements for the keyword \emph{'fracture'} found in the
corpus are: \emph{'Elle' (ullna)}, \emph{'Oberarmknochen'
(humerus)}, \emph{'Sch\"adelgrund' (base of the skull)},
\emph{'Schienbein' (shinbone)}, \emph{'Unterkiefer' (lower jaw)},
\emph{'Unterarmknochen' (radial bone)} etc.

\small
\begin{table}
  \centering
  \caption{Some Complements of a Structure Beginning with
Keyword \emph{'Bruch'} (Fracture).}
  \label{tab-fracture} \scriptsize
  \begin{tabular}{|l|l|c|}
    \hline
    complement & occurrences & top level of GermaNet \\
    \hline
  Rippe & 254 & nomen.koerper \\
  Brustbein & 65 & nomen.koerper \\
  Wirbels\"aule & 58 & nomen.koerper \\
  Sch\"adeldach & 43 &  -- \\
  Oberschenkelknochen & 37 & -- \\
  Sch\"adelbasis & 34 & -- \\
  Schl\"usselbein & 33 & -- \\
  Schambein & 30 & nomen.koerper \\
  Brustwirbels\"aule & 28 & -- \\
  Halswirbels\"aule & 26 & -- \\
  Schulterblatt & 23 & nomen.koerper \\
  \hline
  \end{tabular}
\end{table}
\normalsize

At first, structures with high occurrence frequencies in the
corpus are selected. For this task, the \emph{findings} sections
of the documents are parsed with the syntactic parser of XDOC. A
domain specific grammar with ca. 40 rules is used. In the results
of 18008 parsed sentences, 2808 complex noun phrases \sloppy
(\texttt{NP $\rightarrow$ NP NP$_{genitive}$}) with 1069 different
keywords are encountered.

The most frequent keywords in such structures are: \emph{'Abgang'
(outlet)}, \emph{'Bauchteil' (abdominal part)}, \emph{'Brustteil'
(chest part)}, \emph{'Blutreichtum' (hyperemia)},
\emph{'F\"aulnis' (sepsis)}, \emph{'Haut' (dermis)},
\emph{'Schleimhaut' (mucosa)}, \emph{'Gegend' (region)},
\emph{'Schnitt\-fl\"achen' (cut surfaces)}, \emph{'Unterblutung'
(hematoma)}, and \emph{'Bruch' (fracture)}.

The next step is to use regular expressions to get all occurrences
of a particular combination of a keyword and a complement, because
not all occurrences from the corpus can be obtained with the chart
parser. The reason for this is that there are gaps in the grammar
(when parsing the section \emph{background} and \emph{discussion})
and gaps in the morphological lexicon.

The most frequent keywords in regular expressions are used to get
all phrases that begin with the keyword. The length of these
phrases (text window size) is restricted to be 3 tokens (or 4
tokens, when adjectives in the complement noun phrase) are
allowed.

For each structure, the GermaNet interface is used to check if
information about the keyword of the complement NP is available.
For the example (keyword: \emph{fracture}), GermaNet contains 31
complement elements of the 93 complement elements found in our
corpus. Most complement words of a keyword found in GermaNet have
the same top level category, only a small number of words have
more than one reading. For the example, following top level
categories (given with its percentage related to all senses) are
encountered: $<$nomen.Koerper$>$: 75 \%, $<$nomen.Artefakt$>$:
16,5 \%, $<$nomen.Menge$>$: 5,5 \%, and $<$nomen.Nahrung$>$: 3 \%.
All the words with more than one sense have at least one sense
with the top level category $<$nomen.Koerper$>$.

Table \ref{tab-fracture} presents a small excerpt of the
complement words\footnote{The complement words
  described in table \ref{tab-fracture} occurred in the corpus in a singular or plural form.}
  in the corpus for the keyword \emph{fracture}.
The main top level category for the complement words is
$<$nomen.Koerper$>$ (WordNet category: noun.body).

The first assumption is that all complement words of a keyword in
a domain will belong to the same top level category in GermaNet.
That means that those words of the example which are not contained
in GermaNet, like \emph{'Oberarmknochen' (humerus)},
\emph{'Sch\"adelbasis' (base of the skull)}, \emph{'Sch\"adeldach'
(calvarium)}, \emph{'Brustwirbels\"aule' (thoracic spine)}, etc.,
can be assigned to the same top level category:
$<$nomen.Koerper$>$. In the case of the example (keyword
\emph{fracture}), this heuristic yields the correct top level
category for 93,44 \% of all complements.

In the next step, subclasses of the GermaNet top level category
will be used , so that a word can be annotated with additional
information, e.g., hypernymy relation. For this task, GermaNet's
hypernymy relation is exploited. The hypernym information for all
complements is selected, which do exist in GermaNet. The hypernymy
relation in GermaNet can contain more than one level of hypernyms
for an entry.

At first, all senses with their hypernym information are selected.
Each sense and its hypernyms describe a class path and each entry
in this class path names a semantic class.  The occurrences of the
different semantic classes for all senses (class paths) are
counted. For the different forms of the phrase \emph{'Bruch
der/des XXX'} (in English: fracture of XXX), 36 senses with
altogether 63 different semantic classes are encountered. Table
\ref{tab-numberfirst} presents a partial list of all semantic
classes and its number of occurrences in all the senses for the
complement elements covered by GermaNet. For example, the semantic
class \emph{'Knochen' (bone)} appears in 13 senses as a hypernym,
the semantic class \emph{'Computerprogramm' (software)} only in
one sense.

\small
\begin{table}
  \centering
  \caption{Hypernym Information for Complement Entries.}\label{tab-numberfirst}
  \scriptsize \begin{tabular}
  {|l|c|c|}\hline
    hypernym & number of & percentage\\
    & occurrences & \\\hline
     $<$nomen.Tops$>=>$ Objekt& 22 & 13.75 \\
    $<$nomen.Koerper$>=>$
Hornsubstanz & 13 & 8.125 \\
    $<$nomen.Substanz$>=>$ Stoff1, Substanz, & & \\
Materie & 13 & 8.125 \\
    $<$nomen.Koerper$>=>$ K\"orpersubstanz & 13 & 8.125 \\
    $<$nomen.Koerper$>=>$ Knochen, Gebein & 13 & 8.125 \\
    $<$nomen.Artefakt$>=>$ Artefakt, Werk & 7 & 4.375 \\
    $<$nomen.Tops$>=>$& &\\
Ding, Sache, Gegenstand, Gebilde & 7 &4.375 \\
    $<$nomen.Menge$>=>$& &\\
Masseinheit, Mass, Messeinheit, Messeinheit*o & 2 & 1.25 \\
    ... & ... & ... \\
$<$nomen.Koerper$>=>$ Armknochen & 2 & 1.25 \\
    $<$nomen.Artefakt$>=>$ Computerprogramm, & &\\
Programm & 1 & 0.625 \\
$<$nomen.Artefakt$>=>$ ?akustisches Ger\"at& 1 &0.625\\ \hline
  \end{tabular}
\end{table}
\normalsize

At this point, we don't have a clear and unique result. The highly
frequent hypernym entries in all senses found in GermaNet are the
entries: \emph{'Objekt'} (object), \emph{'Hornsubstanz'}
(akeratosis), \emph{'Knochen'} (bone), etc. These results can be
enhanced when we allow only senses that describe a concept with
the top level assignment of $<$nomen.Koerper$>$ (see table
\ref{tab-numbersnd}). The possible senses are reduced to 27 senses
with altogether 22 different semantic classes.

When the basic concepts (WordNet's 'unique beginner') of GermaNet,
e.g. \emph{Objekt} is ignored, and when the most specific hypernym
of all high frequent hypernyms is selected, the following partial
class path results:

\scriptsize
\begin{alltt}
<nomen.Koerper>=> Knochen, Gebein
       <nomen.Koerper>=> Hornsubstanz
           <nomen.Koerper>=> K{\"o}rpersubstanz
               <nomen.Substanz>=> Stoff, Substanz,
                    Materie
                   <nomen.Tops>=> Objekt
\end{alltt}
\normalsize

For the selection of the most specific hypernym, every level in
the class path is assigned with a weighting factor (The selection
process can be described by the Eq. \ref{manuOntolex}). The unique
beginner starts with the factor \emph{0} (in our example
\emph{Objekt}), the next higher level get the factor \emph{1}, and
so on.

\begin{equation}\label{manuOntolex}
c_i  = \mathop {\arg \max }\limits_{c_i } \frac{{f_i n(c_i )}} {N}
\end{equation}
\normalsize

\begin{figure}[hbtp]
\begin{center}
\includegraphics[scale=0.4]{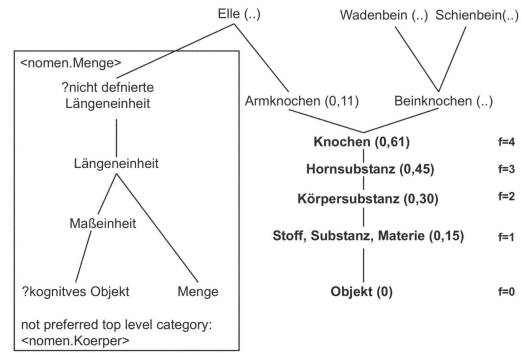}
\caption{Weighting of Possible Semantic Classes.}\label{selection}
\end{center}
\end{figure}

For each semantic class $c_i$, the quotient (occurrences of the
semantic class n($c_i$) divided by number of all semantic classes
N) is multiplied by its weighting factor $f_i$ (see also Fig.
\ref{selection}). In the result above, the semantic classes got
following factor assignment: $f_{Objekt}$ -- 0, $f_{Stoff}$ -- 1,
$f_{Koerpersubstanz}$ -- 2, $f_{Hornsubstanz}$ -- 3, $f_{Knochen}$
-- 4.

\small

\normalsize


The whole approach described above is sketched in the following
(given a keyword \emph{K} and a set of all complements \emph{$C_S$} of \emph{K}):\\
\scriptsize 
\begin{description}
\item \textbf{procedure} find-entry (K, $C_S$):
   \begin{description}
   \item Step 1: \textbf{for each} complement c $\epsilon$ $C_S$: get all (GermaNet) senses of c $\rightarrow$ $H_S$;
   \item Step 2: ascertain the most frequent top level category in $H_S$ $\rightarrow$ T;
   \item Step 3: remove senses from $H_S$, which are not assigned with the preferred top level category T $\rightarrow$
$H_{Sprefer}$;
   \item Step 4: \textbf{for each} sense s $\epsilon$ $H_{Sprefer}$: collect all semantic classes of the hypernym information of s $\rightarrow$
   $SC_S$;
   \item Step 5: \textbf{for each} semantic class sc $\epsilon$ $SC_S$: calculate
            \begin{description}
            \item Step 5.1: occurrences of sc (n($c_{i}$))/number of all sc (N)$\rightarrow$
            $sc_{ratio}$;
            \item Step 5.2: $sc_{ratio}$ times level in the hypernym tree ($f_{i}$) $\rightarrow$
            $sc_{weight}$;
            \end{description}
   \item Step 6: select sc with maximum of $sc_{weight}$;
\end{description}
\end{description}
\normalsize

For ca. 80 \% of the complement words of the keyword
\emph{fracture} this assignment is correct. Erroneous assignments
result from misspelling of tokens (e.g. \emph{Oberschenkelknorren}
instead \emph{Oberschenkelknochen}) or erroneous fragments in the
results of the preprocessing steps (e.g., the treatment of
German's truncations in phrases like \emph{Bruch des Ober- und
Unterarmes (fracture of upper arm and forearm)}). Another type of
error occurring in the evaluation was the case when the second
noun phrase can also be parsed as a complex noun phrase. For the
example, only 2 forms are encountered: \emph{Bruch der Anteile ...
(fracture of parts of ...)} and \emph{Bruch der Wandung ...
(fracture of septum of ...)}. For a reliable evaluation of these
results, it is necessary to consult the domain specific knowledge
of a medical expert. In some cases, for a non-expert it is not
clear if a derived sense is correct. For instance, the word
\emph{'Ellenbogengelenk' (elbow joint)} describes a (complex)
system of bones, cartilages, connective tissues, etc.

\subsection{Compound Analysis}

An alternative way is to group words according to their
components. In German and especially in the corpus, a lot of
compounds are found, e.g., \emph{'Armknochen' (arm bones)},
\emph{'Oberarmknochen' (upper arm bone)}, and
\emph{'Unterarmknochen' (forearm bone)}. GermaNet contains the
word \emph{'Armknochen'}, but not the words
\emph{'Oberarmknochen'} and \emph{'Unterarmknochen'}. For this
case, a list of typical prefixes of the domain can be made of use.
Prefixes in the domain are e.g., \emph{'Unter-'}, \emph{'Ober-'},
\emph{'Innen-'}, \emph{'Aussen-'}, quasi a pair list of antonyms.
In this case, the hypernym information can be used directly for
the new entry. For example, in GermaNet following entry of the
word \emph{Armknochen} is encountered:

\scriptsize
\begin{alltt}
1 sense of armknochen

Sense 1 <nomen.Koerper>Armknochen
       <nomen.Koerper>=> Knochen, Gebein
           <nomen.Koerper>=> Hornsubstanz
               <nomen.Koerper>=> K{\"o}rpersubstanz
                   <nomen.Substanz>=> Stoff1, Substanz,
                        Materie
                       <nomen.Tops>=> Objekt
\end{alltt}
\normalsize

In the corpus,  the complement words \emph{'Unterarmknochen'} (3
times) and \emph{'Oberarmknochen'} (19 times) for the same keyword
are found. Both have no entry in GermaNet. The following
information for the word \emph{'Oberarmknochen'} (similar for the
word \emph{'Unterarmknochen'}) could be inserted:

\scriptsize

\begin{alltt}
 <nomen.Koerper>Oberarmknochen
    <nomen.Koerper>=> Armknochen
       <nomen.Koerper>=> Knochen, Gebein
           <nomen.Koerper>=> Hornsubstanz
               <nomen.Koerper>=> K{\"o}rpersubstanz
                   <nomen.Substanz>=> Stoff1, Substanz,
                        Materie
                       <nomen.Tops>=> Objekt
\end{alltt}
\normalsize

Another kind of compound in the corpus are compounds with a prefix
that describes a \texttt{body part}, e.g.
\emph{'Nierenschleimhaut'} (kidney mucosa),
\emph{'Brustwirbels\"aule'} (thoracic spine). \texttt{body part}
can be named a region of the body or an organ. In this case, the
following restrictions should be considered by the method:
\begin{itemize}
    \item both parts of the compound should have an entry in
    GermaNet and
    \item the parts of the compound should also appear in the corpus as a complex noun phrase: first part of the compound is the
    complement and the second part of the compound should be the
    keyword (e.g., \emph{'Magenschleimhaut'} (stomach mucosa) vs. \emph{'Schleimhaut des Magens'} (mucosa of stomach).
\end{itemize}
\normalsize

In these cases, information via GermaNet's meronym
relation is deduced.

\small
\begin{table}
  \centering
  \caption{Enhanced Hypernym Information for Complement Entries.}\label{tab-numbersnd}
  \scriptsize
  \begin{tabular}{|l|c|c|}\hline
    hypernym & number of & percentage \\
    & occurrences & \\\hline
     $<$nomen.Tops$>=>$ Objekt& 14 & 16.47 \\
    $<$nomen.Koerper$>=>$
Hornsubstanz & 13 & 15.29 \\
    $<$nomen.Substanz$>=>$ Stoff1, Substanz,& &\\
Materie & 13 & 15.29 \\
    $<$nomen.Koerper$>=>$ K\"orpersubstanz & 13 & 15.29 \\
    $<$nomen.Koerper$>=>$ Knochen, Gebein & 13 & 15.29 \\
    $<$nomen.Artefakt$>=>$ Artefakt, Werk & -- & -- \\
    $<$nomen.Tops$>=>$ & &\\
Ding, Sache, Gegenstand, Gebilde & -- & -- \\
    $<$nomen.Menge$>=>$& &\\
Masseinheit, Mass, Messeinheit, Messeinheit*o & -- & -- \\
    ... & ... & ... \\
$<$nomen.Koerper$>=>$ Armknochen & 2 & 2.35 \\
$<$nomen.Artefakt$>=>$ Computerprogramm,& &\\
Programm & -- & -- \\
$<$nomen.Artefakt$>=>$ ?akustisches Ger\"at& -- & --\\ \hline
  \end{tabular}
\end{table}
\normalsize

\subsection{Disambiguation}
The fundament of correct deduction of concepts is the selection of
the correct sense of the senses available in GermaNet. In our
case, the restriction to one top level category is sufficient for
this analysis of forensic autopsy protocols, especially the
findings section. In this section, only anatomic concepts and its
findings are described. For other domains, it is necessary to use
methods for a certain word sense disambiguation, e.g., methods
that used selectional preference ( see \cite{Resnik1997} or
\cite{abney.light.1999}) or conceptual density
(\cite{agirre96word}) for word sense disambiguation.

\section{Related Work}
The approach exploits the specific syntactic structures of a
sublanguage. In the work of
\cite{kokkinakis.toporowska.et.al.2000.lrec2000}, the analyses of
compounds and specific syntactic structures are used for the
extension of the Swedish SIMPLE lexicon. This work exploits the
advantage of the productive compounding characteristic of Swedish
to derive new lexical items (results in information about semantic
type, domain, and semantic class). Furthermore, they used a raw
and partially parsed corpus for the analyses of enumerative NPs
(with more than three common nouns) for the derivation of
co-hyponyms. The following heuristic is used for an unknown noun
in an enumerative NP: if at least two nouns have the same
assignment to a semantic class, then there is a strong indication
that the rest of the nouns are co-hyponyms and thus semantically
similar with the two already encoded nouns.


The usage of a lexical resource to learn new entries for the same
resource (WordNet) is described in \cite{navigli:2002ontolex}.
This paper outlines an approach for the deduction of a sense of
multi-word terms that is based on the senses of individual words
of the multi-word terms. Another similar approach that combines
corpus and WordNet information to deliver verb synonyms for high
frequent verbs of a domain-specific sublanguage is described by
Xiao \cite{xiao.roesner.2004wordnet}. Peters \cite{peters:04}
describes how new knowledge fragments can be derived and extended
from synonymy, hypernymy and thematic relations of WordNet and
implicit information from the (Euro)WordNet.

\section{Conclusion}
Linguistic resources with domain-specific coverage are crucial for
the development of concrete application systems. In this paper, we
proposed an approach for the extraction of semantic information,
using the information available in GermaNet for the individual
words that frequently occur in a specific syntactic structure of
the corpus.

The results of the approach can be helpful for the corpus based
semiautomatic extension of the GermaNet resources. With this
approach, it is possible to extract information about a new entry
(e.g., \emph{forearm bone}) or to complete senses or hypernym
information for entries existing in GermaNet (e.g., \emph{lower
leg}). The results also contain synonyms, like \emph{'Jochbogen'
(zygoma)}, \emph{'Jochbeinknochen' (zygomatic bone)}, and
\emph{'Jochbogen' (zygomatic)}, which can be detected by an deeper
context-related investigation of the elements of a complement set.

In future work, we will evaluate the approach for other syntactic
structures and investigate if it is possible to deduce information
about the keyword of a syntactic structure when the complements
are known. Another aspect will be the exploitation of the
resources of the Medical Subject Headings
(MeSH).\footnote{http://www.nlm.nih.gov/mesh/meshhome.html} The
investigation points are: How many medical terms (in a more
everyday language) of the forensic autopsy protocols are covered
by MeSH? and What differences exist between entries of MeSH and
GermaNet, because Basili et al. describe some discrepancies
between entries in MeSH and WordNet \cite{basili.webintell.2003}.
Further on, this paper outlines the mapping of a domain concept
hierarchy (MeSH) with a lexical knowledge base (WordNet) for the
building of a linguistically motivated domain hierarchy. If such
an approach is necessary in the analysis of forensic autopsy
protocols, it should be considered in further analyses of the
corpus and the evaluation by medical experts.


\bibliographystyle{lrec2000}
\bibliography{xample}

\end{document}